\documentclass[11pt,a4paper]{article}
\usepackage[hyperref]{acl2020}
\usepackage{times}
\usepackage{latexsym}
\usepackage{amsmath}
\usepackage{amssymb}
\usepackage{tabularx}

\usepackage{graphicx}

\usepackage{tikz}

\usetikzlibrary{fit,positioning, matrix,decorations.pathreplacing,calc,fit,backgrounds}

\definecolor{coolgrey}{rgb}{0.55, 0.57, 0.67}
\usepackage{microtype}

\aclfinalcopy % Uncomment this line for the final submission
\title{A Transformer-based joint-encoding for Emotion Recognition and Sentiment Analysis}

 \author{Jean-Benoit Delbrouck \and No\'e Tits \and Mathilde Brousmiche \and St\'ephane Dupont \\
          Information, Signal and Artificial Intelligence Lab\\
          University of Mons, Belgium\\
          \{jean-benoit.delbrouck, noe.tits, mathilde.brousmiche,  stephane.dupont\}@umons.ac.be}

\date{}

\begin{document}
\maketitle
\begin{abstract}
Understanding expressed sentiment and emotions are two crucial factors in human multimodal language. This paper describes a Transformer-based joint-encoding (TBJE) for the task of Emotion Recognition and Sentiment Analysis. In addition to use the Transformer architecture, our approach relies on a modular co-attention and a glimpse layer to jointly encode one or more modalities. The proposed solution has also been submitted to the ACL20: Second Grand-Challenge on Multimodal Language to be evaluated on the CMU-MOSEI dataset. The code to replicate the presented experiments is open-source \footnote{\url{https://github.com/jbdel/MOSEI_UMONS}}.
\end{abstract}

\section{Introduction}

Predicting affective states from multimedia is a challenging task. Emotion recognition task has existed working on different types of signals, typically audio, video and text. Deep Learning techniques allow the development of novel paradigms to use these different signals in one model to leverage joint information extraction from different sources. This paper aims to bring a solution based on ideas taken from Machine Translation (Transformers, \citet{vaswani2017attention}) and Visual Question Answering (Modular co-attention, \citet{yu2019deep}). Our contribution is not only very computationally efficient, it is also a viable solution for Sentiment Analysis and Emotion Recognition. Our results can compare with, and sometimes surpass, the current state-of-the-art for both tasks on the CMU-MOSEI dataset \cite{bagher-zadeh-etal-2018-multimodal}. \\

This paper is structured as follows: first, in section \ref{sec:related}, we quickly go over the related work that have been evaluated on the MOSEI dataset, we then proceed to describe our model in Section \ref{sec:model}, we then explain how we extract our modality features from raw videos in Section \ref{sec:feat_ex} and finally, we present the dataset used for our experiments and their respective results in section \ref{sec:dataset} and \ref{sec:results}.
\section{Related work}
\label{sec:related}

Over the years, many creative solutions have been proposed by the research community in the field of Sentiment Analysis and Emotion Recognition. In this section, we proceed to describe different models that have been evaluated on the CMU-MOSEI dataset. To the best of our knowledge, none of these ideas uses a Tansformer-based solution. \\

The Memory Fusion Network (MFN, \citet{zadeh2018memory}) synchronizes multimodal sequences
using a multi-view gated memory that stores intraview and cross-view interactions through time. \\

Graph-MFN \cite{bagher-zadeh-etal-2018-multimodal} consists of a Dynamic Fusion Graph (DFG) built upon MFN. DFG is a fusion technique that tackles the nature of cross-modal dynamics in multimodal language. The fusion is a network that learns to models the n-modal interactions and can dynamically alter its structure to choose the proper fusion graph based on the importance of each n-modal dynamics during inference. \\

\citet{sahay-etal-2018-multimodal} use Tensor Fusion Network (TFN), i.e.  an outer product of the modalities. This operation can be performed either on a whole sequence or frame by frame. The first one lead to an exponential increase of the feature space when modalities are added that is computationally ex-pensive. The second approach was thus preferred. They showed an improvement over an early fusion baseline. \\

Recently, \citet{shenoy2020multiloguenet} propose a solution based on a context-aware RNN, Multilogue-Net, for Multi-modal Emotion Detection and Sentiment Analysis in conversation.

\section{Model}
\label{sec:model}
This section aims to describe the two model variants evaluated in our experiment: a monomodal variant and a multimodal variant. The monomodal variant is used to classify emotions and sentiments based solely on L (Linguistic), on V (Visual) or on A (Acoustic). The multimodal version is used for any combination of modalities. \\

Our model is based on the Transformer model \cite{vaswani2017attention}, a new encoding architecture that fully eschews recurrence for sequence encoding and instead relies entirely on an attention mechanism and Feed-Forward Neural Networks (FFN) to draw global dependencies between input and output. The Transformer allows for significantly more parallelization compared to the Recurrent Neural Network (RNN) that generates a sequence of hidden states $h_t$, as a function of the previous hidden state $h_{t-1}$ and the input for position $t$. \\

\subsection{Monomodal Transformer Encoding}
\label{sec:mono}
The monomodal encoder is composed of a stack of $B$ identical blocks but with their own set of training parameters. Each block has two
sub-layers. There is a residual connection around each of the two sub-layers, followed by layer normalization \cite{ba2016layer}. The output of each sub-layer can be written like this:
\begin{equation}
    \text{LayerNorm}(x + \text{Sublayer}(x))
\end{equation}
where Sublayer(x) is the function implemented by the sub-layer itself. In traditional Transformers, the two sub-layers are respectively a multi-head self-attention mechanism and a simple Multi-Layer Perceptron (MLP).

The attention mechanism consists of a Key $K$ and Query $Q$ that interacts together to output a attention map applied to Context C:
\begin{equation}
    \text{Attention}(Q, K, C) = \text{softmax}(\frac{QK^\top}{\sqrt{k}})C
\end{equation}
% We suspect that for large values of dk, the dot products grow large in magnitude, pushing the softmax function into regions where it has extremely small gradients 4.
In the case of self-attention, $K$, $Q$ and $C$ are the same input. If this input is of size $N \times k$, the operation $QK^\top$ results in a squared attention matrix containing the affinity between each row $N$. Expression $\sqrt{k}$ is a scaling factor. The multi-head attention (MHA) is the idea of stacking several self-attention attending the information from different representation sub-spaces at different positions:

\begin{equation}
    \begin{aligned}
    \text{MHA}(Q, K,C) = \text{Concat}(\text{head}_1, ..., \text{head}_h) W_o \\
    \text{where}\;\text{head}_i = \text{Attention}(QW_i^Q, KW_i^K, CW_i^C)
    \end{aligned}
\end{equation}

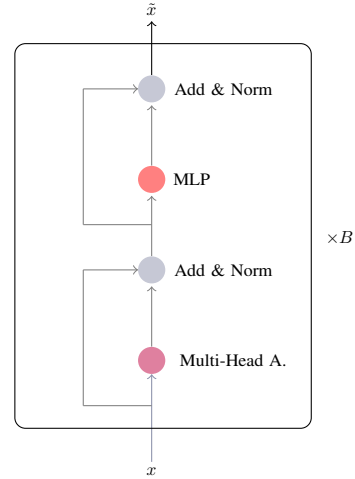
\begin{figure}[h]
\centering
\begin{tikzpicture}[scale=0.6, transform shape]
    \tikzstyle{every pin edge}=[<-,shorten <=1pt]
    \tikzstyle{neuron}=[circle,fill=black!25,minimum size=17pt,inner sep=0pt]
    \tikzstyle{input neuron}=[neuron, fill=green!50];
    \tikzstyle{func neuron}=[neuron, fill=purple!50];
    \tikzstyle{att neuron}=[neuron, fill=coralpink!50];
    \tikzstyle{fake neuron}=[neuron, fill=white!50];
    \tikzstyle{fusion neuron}=[neuron, fill=coolgrey!50];
    \tikzstyle{hidden neuron}=[neuron, fill=red!50];

    \tikzstyle{annot} = [text width=4em, text centered]

    \node[func neuron] (A) at (0,0) {};
    \node[annot,below of=A, node distance=70pt] (xin) {$x$};
    \path[->,draw=coolgrey] (xin) -- (A);  
    \node[fusion neuron] (B) at (0,2) {};
    \node[hidden neuron] (C) at (0,4) {};
    \node[fusion neuron] (D) at (0,6) {};

    \path[shorten >=1pt,->,draw=black!50] (A) edge (B);
    \path[shorten >=1pt,->,draw=black!50] (B) edge (C);
    \path[shorten >=1pt,->,draw=black!50] (C) edge (D);

    \draw[rounded corners] (-3, -1.5) rectangle (3.5,7) {};
    \node[annot,above right = 0.2cm and 3cm of B] () {$\times B$};

    \node[annot,below left = 1.5cm and -0.8cm of A] () {};
    \node[annot,right of=A, node distance=50pt,text width=10em] () {Multi-Head A.};
    
    \draw[draw=black!50] (0, -1) -- (-1.5, -1);
    \path[draw=black!50] (-1.5, -1) -- (-1.5, 2);
    \path[shorten >=0pt,->,draw=black!50] (-1.5, 2) -- (-0.3,2);

    \node[annot,right of=B, node distance=45pt,text width=10em] () {Add \& Norm};

    \draw[draw=black!50] (0, 3) -- (-1.5, 3);
    \path[draw=black!50] (-1.5, 3) -- (-1.5, 6);
    \path[shorten >=0pt,->,draw=black!50] (-1.5, 6) -- (-0.3,6);

    \node[annot,right of=C, node distance=25pt] () {{MLP}};
    \node[annot,right of=D, node distance=45pt,text width=10em] () {Add \& Norm};
    
    \node[annot,above of=D, node distance=50pt] (xout) {$\tilde{x}$};
    \path[->,draw=black] (D) -- (xout);

\end{tikzpicture}
\caption{Monomodal Transformer encoder.}
\label{fig:transformer}
\end{figure}

A subspace is defined as slice of the feature dimension $k$. In the case of four heads, a slice would be of size $\frac{k}{4}$. The idea is to produce different sets of attention weights for different feature sub-spaces.
After encoding through the blocks, output $\tilde{x}$ can be used by a projection layer for classification. In Figure \ref{fig:transformer}, $x$ can be any modality feature as described in Section \ref{sec:feat_ex}.

\subsection{Multimodal Transformer Encoding}
\label{sec:multimo}
The idea of a multimodal transformer consists in adding a dedicated transformer (section \ref{sec:mono}) for each modality we work with. While our contribution follows this procedure, we also propose three ideas to enhance it: a joint-encoding, a modular co-attention \cite{yu2019deep} and a glimpse layer at the end of each block. \\

The modular co-attention consists of modulating the self-attention of a modality, let's call it $y$, by a primary modality $x$. To do so, we switch the key $K$ and context $C$ of the self-attention from $y$ to $x$. The operation $QK^\top$ results in an attention map that acts like an affinity matrix between the rows of modality matrix $x$ and  $y$. This computed alignment is applied over the context $C$ (now $x$) and finally we add the residual connection $y$. The following equation describes the new attention sub-layer:

\begin{equation}
   y = \text{LayerNorm}(y + \text{MHA}(y,x,x)) \label{eq:block_output}
\end{equation}

In this scenario, for the operation $QK^\top$ to work as well as the residual connection (the addition), the feature sizes of $x$ and $y$ must be equal. This can be adjusted with the different transformation matrices of the MHA module. Because the encoding is joint, each modality is encoded at the same time (i.e. we don't unroll the encoding blocks for one modality before moving on to another modality). This way, the MHA attention of modality $y$ for block $b$ is done by the representation of $x$ at block $b$. \\

Finally, we add a last layer at the end of each modality block, called the glimpse layer, where the modality is projected in a new space of representation. A glimpse layer consists of stacking $G$ soft attention layers and stacking their outputs. Each soft attention is seen as a glimpse. Formally, we define the soft attention (SoA) $i$ with input matrix $M \in \mathbb{R}^{N \times k}$ by a MLP and a weighted sum:

\begin{equation}
    \begin{aligned}
        a_i &= \text{softmax}({v_i^a}^\top (W_m M)) \\
    \text{SoA}_i(M) = m_i &= \sum\limits_{j=0}^{N} a_{ij}M_j
    \end{aligned}
\end{equation}
where $W_m$ if a transformation matrix of size $2k \times k$, $v_i^a$ is of size $1 \times 2k$ and $m_i$ a vector of size $k$. Then we can define the glimpse mechanism for matrix $M$ of glimpse size $G^m$ as the stacking of all glimpses:

\begin{equation}
    \text{G}_{M} = \text{Stacking}({m_1}, \hdots, m_{G^m}) \nonumber  \\
\end{equation}

Note that before the parameter $W_m$, whose role is to embed the matrix $M$ in a higher dimension, is shared between all glimpses (this operation is therefore only computed once) while the set of vectors $\{v_i^a\}$ computing the attention weights from this bigger space is dedicated for each glimpse. In our contribution, we always chose $G^m$ = $N$ so the sizes allow us to perform a final residual connections $M = \text{LayerNorm}(M + \text{G}_{M}$). \\

\begin{figure}[h]
\centering
\begin{tikzpicture}[scale=0.51, transform shape]
    \tikzstyle{every pin edge}=[<-,shorten <=1pt]
    \tikzstyle{neuron}=[circle,fill=black!25,minimum size=17pt,inner sep=0pt]
    \tikzstyle{input neuron}=[neuron, fill=green!50];
    \tikzstyle{func neuron}=[neuron, fill=purple!50];
    \tikzstyle{att neuron}=[neuron, fill=coralpink!50];
    \tikzstyle{fake neuron}=[neuron, fill=white!50];
    \tikzstyle{fusion neuron}=[neuron, fill=coolgrey!50];
    \tikzstyle{hidden neuron}=[neuron, fill=red!50];
    \tikzstyle{glimpse neuron}=[neuron, fill=cyan!50];

    \tikzstyle{annot} = [text width=4em, text centered]

    \node[func neuron] (A) at (0,0.7) {};
    \node[annot,below of=A, node distance=90pt] (xin) {$x$};
    \path[->,draw=coolgrey] (xin) -- (A);  
    
    \node[fusion neuron] (B) at (0,2) {};
    \node[hidden neuron] (C) at (0,4) {};
    \node[fusion neuron] (D) at (0,6) {};
    \node[glimpse neuron] (E) at (0,8) {};
    \node[fusion neuron] (F) at (0,10) {};
    
    \path[shorten >=1pt,->,draw=black!50] (A) edge (B);
    \path[shorten >=1pt,->,draw=black!50] (B) edge (C);
    \path[shorten >=1pt,->,draw=black!50] (C) edge (D);
    \path[shorten >=1pt,->,draw=black!50] (D) edge (E);
    \path[shorten >=1pt,->,draw=black!50] (E) edge (F);

    \draw[rounded corners] (-2.5, -1.5) rectangle (3.5,11) {};
    \node[annot,above left = 2.2cm and 2cm of B] () {$\times B$};

    % \node[annot,below left = 1.5cm and -0.8cm of A] () {$E^x$};
    \node[annot,right of=A, node distance=50pt,text width=10em] () {Multi-Head A.};
    
    \draw[draw=black!50] (0, -1) -- (-1.5, -1);
    \path[draw=black!50] (-1.5, -1) -- (-1.5, 2);
    \path[shorten >=0pt,->,draw=black!50] (-1.5, 2) -- (-0.3,2);

    \node[annot,right of=B, node distance=45pt,text width=10em] () {Add \& Norm};

    \draw[draw=black!50] (0, 3) -- (-1.5, 3);
    \path[draw=black!50] (-1.5, 3) -- (-1.5, 6);
    \path[shorten >=0pt,->,draw=black!50] (-1.5, 6) -- (-0.3,6);

    \node[annot,right of=C, node distance=25pt] () {{MLP}};
    \node[annot,right of=D, node distance=45pt,text width=10em] () {Add \& Norm};
    
     \draw[draw=black!50] (0, 7) -- (-1.5, 7);
    \path[draw=black!50] (-1.5, 7) -- (-1.5, 10);
    \path[shorten >=0pt,->,draw=black!50] (-1.5, 10) -- (-0.3,10);

    \node[annot,right of=E, node distance=35pt] () {{Glimpse}};
    \node[annot,right of=F, node distance=45pt,text width=10em] () {Add \& Norm};

    \node[annot,above of=F, node distance=50pt] (xout) {$\tilde{x}$};
    \path[->,draw=black] (F) -- (xout);

    %%%%%%%%%%%%%%%%%%%%%%%%%%%%%%%%%%%
    
%   \node[input neuron, pin=below:$y$] (I) at (0+8,-2) {};
    \node[func neuron] (A) at (0+6.5,0) {};

    \node[annot,below of=A, node distance=70pt] (xin) {$y$};
    \path[->,draw=coolgrey] (xin) -- (A);  
    
    \node[fusion neuron] (B) at (0+6.5,2) {};
    \node[hidden neuron] (C) at (0+6.5,4) {};
    \node[fusion neuron] (D) at (0+6.5,6) {};
    \node[glimpse neuron] (E) at (0+6.5,8) {};
    \node[fusion neuron] (F) at (0+6.5,10) {};

    \path[shorten >=1pt,->,draw=black!50] (A) edge (B);
    \path[shorten >=1pt,->,draw=black!50] (B) edge (C);
    \path[shorten >=1pt,->,draw=black!50] (C) edge (D);
    \path[shorten >=1pt,->,draw=black!50] (D) edge (E);
    \path[shorten >=1pt,->,draw=black!50] (E) edge (F);

    \draw[rounded corners] (-2.5+6.5, -1.5) rectangle (3.5+6.5 , 11) {};
    \node[annot,above right = 2.2cm and 3cm of B] () {$\times B$};

    % \node[annot,below left = 1.5cm and -0.8cm of A] () {$E^x$};
    \node[annot,right of=A, node distance=50pt,text width=10em] () {Multi-Head A.};
    
    \draw[draw=black!50] (0+6.5, -1) -- (-1.5+6.5, -1);
    \path[draw=black!50] (-1.5+6.5, -1) -- (-1.5+6.5, 2);
    \path[shorten >=0pt,->,draw=black!50] (-1.5+6.5, 2) -- (-0.3+6.5,2);

    \node[annot,right of=B, node distance=45pt,text width=10em] () {Add \& Norm};

    \draw[draw=black!50] (0+6.5, 3) -- (-1.5+6.5, 3);
    \path[draw=black!50] (-1.5+6.5, 3) -- (-1.5+6.5, 6);
    \path[shorten >=0pt,->,draw=black!50] (-1.5+6.5, 6) -- (-0.3+6.5,6);

    \node[annot,right of=C, node distance=25pt] () {{MLP}};
    \node[annot,right of=D, node distance=45pt,text width=10em] () {Add \& Norm};
    
    \draw[draw=black!50] (0+6.5, 7) -- (-1.5+6.5, 7);
    \path[draw=black!50] (-1.5+6.5, 7) -- (-1.5+6.5, 10);
    \path[shorten >=0pt,->,draw=black!50] (-1.5+6.5, 10) -- (-0.3+6.5,10);

    \node[annot,right of=E, node distance=35pt] () {{Glimpse}};
    \node[annot,right of=F, node distance=45pt,text width=10em] () {Add \& Norm};

    \node[annot,above of=F, node distance=50pt] (xout) {$\tilde{y}$};
    \path[->,draw=black] (F) -- (xout);

%%%%%%

    \draw[->,draw=blue] (0.0, -0.1) -- (6.0, -0.1);
    % \draw[draw=blue] (4.5, -1) -- (4.5, 0);
    % \draw[draw=blue] (0.0,-1) -- (4.5, -1);

\end{tikzpicture}
\caption{Multimodal Transformer Encoder for two modalities with joint-encoding.}
\label{fig:transfo_multimodal}
\end{figure}
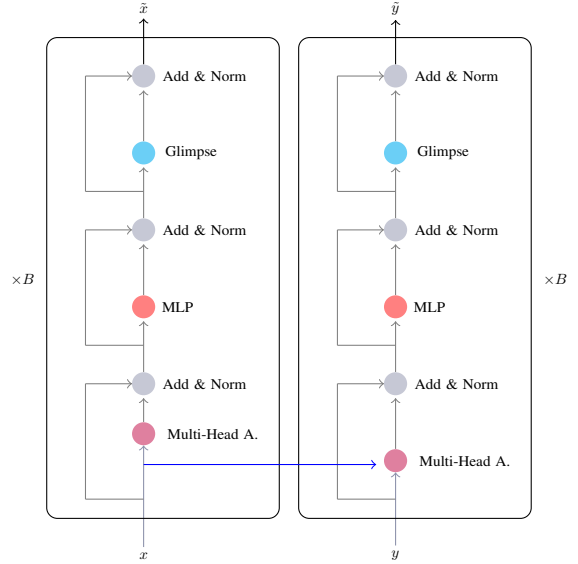

The Figure \ref{fig:transfo_multimodal} depicts the encoding for two features where modality $x$ is modulating the modality $y$. This encoding can be ported to any number of modalities by duplicating the architecture. In our case, it is always the linguistic modality that modulates the others.\\

\subsection{Classification layer}

After all the Transformer blocks were computed, a modality goes into a final glimpse layer of size 1. The result is therefore only one vector. The vectors of each modality are summed element-wise, let's call the results of this sum $s$, and are then projected over possible answers according to the following equation:

\begin{equation}
    y \sim p = W_a (\text{LayerNorm}(s)) \label{eq:proj_layer}
\end{equation}

If there is only one modality, the sum operation is omitted.

\section{Feature extractions}
\label{sec:feat_ex}
This section aims to explain how we pre-compute the features for each modality. These features are the inputs of the Transformer blocks. Note that the features extraction is done independently for each example of the dataset.

\subsection{Linguistic}
Each utterance is tokenized and lowercase. We also remove special characters and punctuation. We build our vocabulary against the train-set and end up with a glossary of 14.176 unique words. We embed each word in a vector of 300 dimensions using GloVe \cite{pennington2014glove}. If a word from the validation or test-set is not in present our vocabulary, we replace it with the unknown token "unk".

\subsection{Acoustic}

% NT: j'ai pas mal broder, donc si jamais à tout hasard, on a trop, on peut virer des truc ici.
% Sinon, n'hésitez pas à améliorer la prose non plus

The acoustic part of the signal of the video contains a  lot of speech. Speech is used in conversations to communicate information with words but also contains a lot of information that are non linguistic such as nonverbal expressions (laughs, breaths, sighs) and prosody features (intonation, speaking rate). These are important data in an emotion recognition task. \\

% Many different acoustic feature sets were designed to extract relevant information from speech. 

Acoustic features widely used in the speech processing field such as F0, formants, MFCCs, spectral slopes consist of handcrafted sets of high-level features that are useful when an interpretation is needed, but generally discard a lot of information. Instead, we decide to use low-level features for speech recognition and synthesis, the mel-spectrograms. Since the breakthrough of deep learning systems, the mel-spectrograms have become a suitable choice. \\

%Mel are based on 
The spectrum of a signal is obtained with Fourier analysis that decompose a signal in a sum of sinusoids. The amplitudes of the sinusoids constitute the amplitude spectrum. A spectrogram is the concatenation over time of spectra of windows of the signal. Mel-spectrogram is a compressed version of spectrograms, using the fact the human ear is more sensitive to low frequencies than high frequencies. This representation thus attributes more resolution for low frequencies than high frequencies using mel filter banks. A mel-spectrogram is typically used as an intermediate step for text-to-speech synthesis~\cite{dctts-18-tachibana} in state-of-the-art systems as audio representation, so we believe it is a good compromise between dimensionality and representation capacity. \\

Our mel-spectrograms were extracted with the same procedure as in~\cite{dctts-18-tachibana} with librosa~\cite{librosa-15-mcfee} library with 80 filter banks (the embedding size is therefore 80). A temporal reduction by selecting one frame every 16 frames was the applied.

%In this work, we opted for mel-spectrogram from which all this information could be computed but also contain also the rest of acoustic information.

\subsection{Visual}
Inspired by the success of convolutional neural networks (CNNs) in different tasks, we chose to extract visual features with a pre-trained CNN. Current models for video classification use CNNs with 3D convolutional kernels to process the temporal information of the video together with spatial information \cite{tran2015learning}. The 3D CNNs learn spatio-temporal features but are much more expensive than 2D CNNs and prone to overfitting. To reduce complexity, \citet{tran2018closer} explicitly factorizes 3D convolution into two separate and successive operations, a 2D spatial convolution and a 1D temporal convolution. We chose this model, named R(2+1)D-152, to extract video features for the emotion recognition task. The model is pretrained on Sports-1M and Kinetics. \\

The model takes as input a clip of 32 RGB frames of the video.  Each frame is scaled to the size of 128 x 171 and then cropped a window of size 112 x 112. The features are extracted by taking the output of the spatiotemporal pooling. The feature vector for the entire video is obtained by sliding a window of 32 RGB frames with a stride of 8 frames. \\

We chose not to crop out the face region of the video and keep the entire image as input to the network. Indeed, the video is already centered on the person and we expect that the movement of the body such as the hands can be a good indicator for the emotion recognition and sentiment analysis tasks.

	\begin{table*}[t]
		\centering
		\resizebox{\textwidth}{!}{
		\begin{tabular}{lcccccccccccccc}
		    Test set & \multicolumn{2}{c}{\bf Sentiment}  &\multicolumn{12}{c}{\bf Emotions} \\
			& 2-class & 7-class & \multicolumn{2}{c}{Happy} & \multicolumn{2}{c}{Sad} & \multicolumn{2}{c}{Angry} & \multicolumn{2}{c}{Fear} & \multicolumn{2}{c}{Disgust} & \multicolumn{2}{c}{Surprise} \\
			& $A$ & $A$ & $A$ & F1  & $A$ & F1 & $A$ & F1 & $A$ & F1 & $A$ & F1 & $A$ & F1  \\
			\hline 
			L+ A + V & 81.5 & 44.4 & 65.0 & 64.0 & 72.0 & 67.9 & 81.6 & 74.7 & 89.1 & 84.0 & 85.9 & 83.6 & 90.5 & 86.1 \\
			L + A & \textbf{82.4} & \textbf{45.5} & \textbf{66.0} & 65.5 & \textbf{73.9} & 67.9 & \textbf{81.9} & 76.0 & \textbf{89.2} & \textbf{87.2} & \textbf{86.5}& 84.5 & \textbf{90.6} & \textbf{86.1} \\
			L & 81.9 & 44.2 & 64.5 & 63.4 & 72.9 & 65.8 & 81.4 & 75.3 & 89.1 & 84.0 & 86.6 & 84.5 & 90.5 & 81.4 \\
			Mu-Net & 82.1 & - & - & \textbf{68.4} & - & \textbf{74.5} & - & \textbf{80.9} & - & 87.0 & - & \textbf{87.3} & - & 80.9 \\
			G-MFN & 76.9 & 45.0 & - & 66.3 & - & 66.9 & - & 72.8 & - & 89.9 & - & 76.6 & - & 85.5
		\end{tabular}}	 
		\caption{Results on the test-set. Note that the F1-scores for emotions are weighted to be consistent with the previous state-of-the-art. Also, we do not compare accuracies for emotions, as previous works use a weighted variant while we use standard accuracy. G-MFN is the Graph-MFN model and  Mu-Net is the Multilogue-Net model.}
        \label{tab:score-tabular}
	\end{table*}%

\section{Dataset}
\label{sec:dataset}
We test our joint-encoding solution on a novel dataset for multimodal sentiment and emotion recognition called CMU-Multimodal Opinion Sentiment and Emotion Intensity (CMU-MOSEI, \citet{bagher-zadeh-etal-2018-multimodal}). It consists of 23,453 annotated sentences from 1000 distinct speakers. Each sentence is annotated for sentiment on a [-3,3] scale from highly negative (-3) to highly positive (+3) and for emotion by 6 classes : happiness, sadness, anger,
fear, disgust, surprise. In the scope of our experiment, the emotions are either present or not present (binary classification), but two emotions can be present at the same time, making it a multi-label problem. \\

\begin{figure}[!ht] 
\centering
\includegraphics[width=\linewidth]{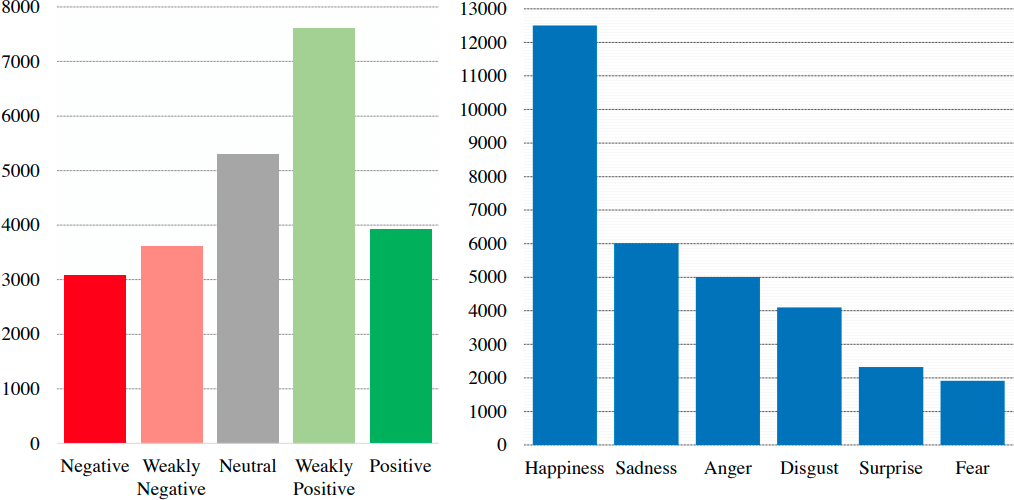}
\caption{MOSEI statistics, taken from the author's paper.}
\label{fig:moseistat}
\end{figure}

The Figure \ref{fig:moseistat} shows the distribution of sentiment and emotions in CMU-MOSEI dataset. The distribution shows a natural skew towards more frequently used emotions. The most common category is happiness with more than 12,000 positive sample points. The least prevalent emotion is fear with almost 1900 positive sample. It also shows a slight shift in favor of positive sentiment. 

\section{Experiments}
\label{sec:results}

In this section, we report the results of our model variants described in Section \ref{sec:model}. We first explain our experimental setting.

\subsection{Experimental settings}
We train our models using the Adam optimizer \cite{kingma2014adam} with a learning rate of $1e-4$ and a mini-batch size of 32. If the accuracy score on the validation set does not increase for a given epoch, we apply a learning-rate decay of factor $0.2$. We decay our learning rate up to 2 times. Afterwards, we use an early-stop of 3 epochs. Results presented in this paper are from the averaged predictions of 5 models. \\

Unless stated otherwise, we use 6 Transformer blocks of hidden-size 512, regardless of the modality encoded. The self-attention has 4 multi-heads and the MLP has one hidden layer of 1024. We apply dropout of 0.1 on the output of each block (equation \ref{eq:block_output}) and of 0.5 on the input of the classification layer ($s$ in equation \ref{eq:proj_layer}). \\

% Our model contains around 40M parameters and converges after 5 epochs. 

\begin{figure}[!ht] 
\centering
\includegraphics[width=\linewidth]{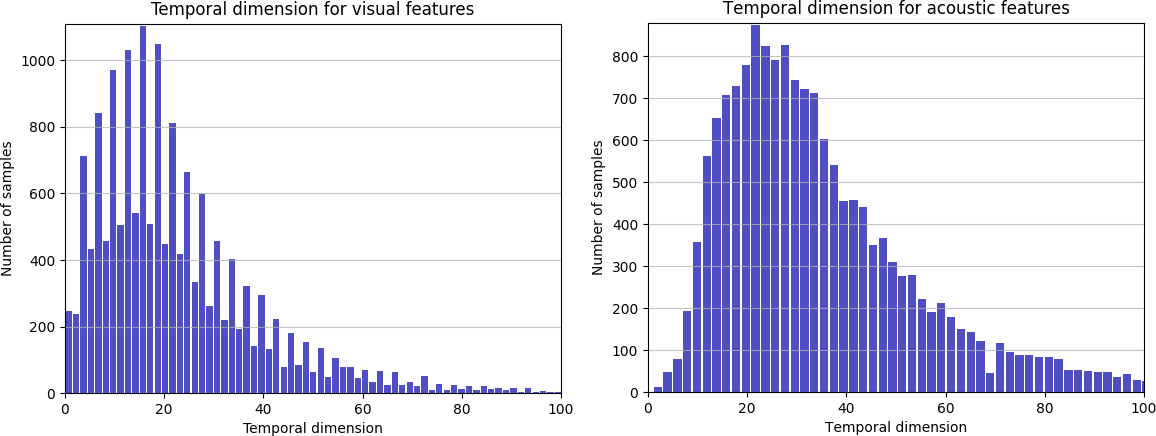}
\caption{Temporal dimension (i.e. rows in our feature matrices) for the acoustic and visual modality.}
\label{fig:tempdim}
\end{figure}

For the acoustic and visual features, we truncate the features for spatial dimensions above 40. We also use that number for the number of glimpses. This choice is made base on Figure \ref{fig:tempdim}

\subsection{Results}
The Table \ref{tab:score-tabular} show the scores of our different modality combinations. We do not compare accuracies for emotions with previous works as they used a weighted accuracy variant while we use standard accuracy. \\

We notice that our L+A (linguistic + acoustic) is the best model. Unfortunately, adding the visual input did not increase the results, showing that it is still the most difficult modality to integrate into a multimodal pipeline. For the sentiment task, the improvement is more tangible for the 7-class, showing that our L+A model learns better representations for more complex classification problems compared to our monomodal model L using only the linguistic input. We also surpass the previous state-of-the-art for this task. For the emotions, we can see that Multilogue-Net gives better prediction for some classes, such as happy, sad, angry and disgust. We postulate that this is because Multilogue is a context-aware method while our model does not take into account the previous or next sentence to predict the current utterance. This might affect our accuracy and f1-score on the emotion task. \\

The following Table \ref{table:private_fold} depicts the results of our solution sent to the Second Grand-Challenge on Multimodal Language. It has been evaluated on the private test-fold released for the challenge and can serve as a baseline for future research. Note that in this table, the F1-scores are unweighted, as should be future results for a fair comparison and interpretation of the results. 

\begin{table}[ht]
	\centering
	\begin{tabular}{lcccccc}
	Sentiment  & 7-class \\
    \hline
    L + A ($A$) & 40.20 \\
    &\\
    Emotion & Happy & Sad & Angry \\
        \hline
    L + A ($A$) & 67.07 & 82.66 & 81.65 \\
    L + A (F1) & 78.08 & 31.42 & 28.38 \\
        &\\
    Emotion & Fear & Disgust & Surprise \\
        \hline
    L + A ($A$) &  88.19 & 79.14 & 90.45 \\
    L + A (F1) &  26.66 & 25.49 & 15.82 \\
	\end{tabular}
\caption{Results on the private test-fold for 7-class sentiment problem and for each emotion. Accuracy is denoted by $A$. In this table, the F1-scores are unweighted, unlike Table \ref{tab:score-tabular}.}
\label{table:private_fold}
\end{table}

\begin{figure}[!ht] 
\centering
\includegraphics[width=\linewidth]{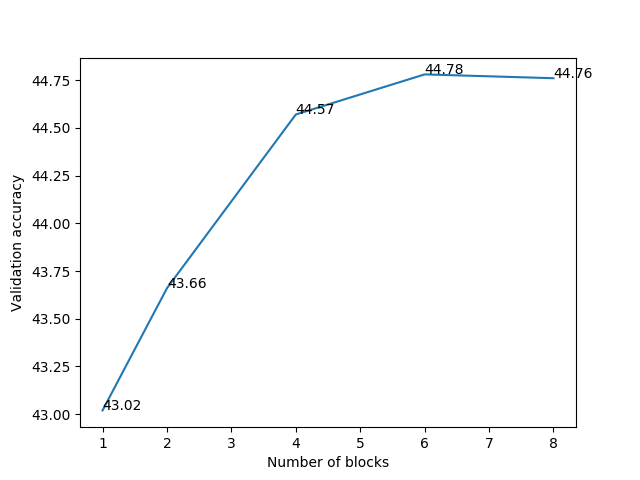}
\caption{7-class sentiment accuracy according to the number of blocks per Transformer.}
\label{fig:blocks}
\end{figure}

\section{Discussions}
We presented a computationally efficient and robust model for Sentiment Analysis and Emotion Recognition evaluated on CMU-MOSEI. Though we showed strong results on accuracy, we can see that there is still a lot of room for improvement on the F1-scores, especially for the emotion classes that are less present in the dataset. To the best of our knowledge, the results presented by our transformer-based joint-encoding are the strongest scores for the sentiment task on the dataset. \\

The following list identifies other features we computed as input for our model that lead to weaker performances:
\begin{itemize}
    \item We tried the OpenFace 2.0 features \cite{baltrusaitis2018openface}. This strategy computes facial landmark, the features are specialized for facial behavior analysis;
    \item We tried a simple 2D CNN named DenseNet \cite{huang2017densely}. For each frame of the video, a feature vector is extracted by taking the output of the average pooling layer;
    \item We tried different values for the number of mel filter bank (512 and 1024) and temporal reduction (1, 2, 4 and 8 frames), we also tried to use the full spectrogram;
    \item We tried not using the GloVe embedding.
\end{itemize}

\section{Acknowledgements}

No\'e Tits  is funded through a FRIA grant (Fonds pour la Formation \`a la Recherche dans l'Industrie et l'Agriculture, Belgium). 

%\newpage
\bibliography{refs}

\begin{thebibliography}{15}
\expandafter\ifx\csname natexlab\endcsname\relax\def\natexlab#1{#1}\fi

\bibitem[{Ba et~al.(2016)Ba, Kiros, and Hinton}]{ba2016layer}
Jimmy~Lei Ba, Jamie~Ryan Kiros, and Geoffrey~E Hinton. 2016.
\newblock Layer normalization.
\newblock \emph{arXiv preprint arXiv:1607.06450}.

\bibitem[{Baltrusaitis et~al.(2018)Baltrusaitis, Zadeh, Lim, and
  Morency}]{baltrusaitis2018openface}
Tadas Baltrusaitis, Amir Zadeh, Yao~Chong Lim, and Louis-Philippe Morency.
  2018.
\newblock {Openface 2.0: Facial behavior analysis toolkit}.
\newblock In \emph{13th IEEE International Conference on Automatic Face \&
  Gesture Recognition}, pages 59--66. IEEE.

\bibitem[{Huang et~al.(2017)Huang, Liu, Van Der~Maaten, and
  Weinberger}]{huang2017densely}
Gao Huang, Zhuang Liu, Laurens Van Der~Maaten, and Kilian~Q Weinberger. 2017.
\newblock Densely connected convolutional networks.
\newblock In \emph{Proceedings of the IEEE Conference on Computer Vision and
  Pattern Recognition}, pages 4700--4708.

\bibitem[{Kingma and Ba(2014)}]{kingma2014adam}
Diederik~P Kingma and Jimmy Ba. 2014.
\newblock Adam: A method for stochastic optimization.
\newblock \emph{arXiv preprint arXiv:1412.6980}.

\bibitem[{McFee et~al.(2015)McFee, Raffel, Liang, Ellis, McVicar, Battenberg,
  and Nieto}]{librosa-15-mcfee}
Brian McFee, Colin Raffel, Dawen Liang, Daniel~PW Ellis, Matt McVicar, Eric
  Battenberg, and Oriol Nieto. 2015.
\newblock librosa: Audio and music signal analysis in python.
\newblock In \emph{Proceedings of the 14th python in science conference}, pages
  18--25.

\bibitem[{Pennington et~al.(2014)Pennington, Socher, and
  Manning}]{pennington2014glove}
Jeffrey Pennington, Richard Socher, and Christopher~D Manning. 2014.
\newblock Glove: Global vectors for word representation.
\newblock In \emph{EMNLP}, volume~14, pages 1532--1543.

\bibitem[{Sahay et~al.(2018)Sahay, Kumar, Xia, Huang, and
  Nachman}]{sahay-etal-2018-multimodal}
Saurav Sahay, Shachi~H Kumar, Rui Xia, Jonathan Huang, and Lama Nachman. 2018.
\newblock \href {https://doi.org/10.18653/v1/W18-3303} {Multimodal relational
  tensor network for sentiment and emotion classification}.
\newblock In \emph{Proceedings of Grand Challenge and Workshop on Human
  Multimodal Language (Challenge-{HML})}, pages 20--27, Melbourne, Australia.
  Association for Computational Linguistics.

\bibitem[{Shenoy and Sardana(2020)}]{shenoy2020multiloguenet}
Aman Shenoy and Ashish Sardana. 2020.
\newblock Multilogue-net: A context aware rnn for multi-modal emotion detection
  and sentiment analysis in conversation.

\bibitem[{Tachibana et~al.(2018)Tachibana, Uenoyama, and
  Aihara}]{dctts-18-tachibana}
Hideyuki Tachibana, Katsuya Uenoyama, and Shunsuke Aihara. 2018.
\newblock Efficiently trainable text-to-speech system based on deep
  convolutional networks with guided attention.
\newblock In \emph{2018 IEEE International Conference on Acoustics, Speech and
  Signal Processing (ICASSP)}, pages 4784--4788. IEEE.

\bibitem[{Tran et~al.(2015)Tran, Bourdev, Fergus, Torresani, and
  Paluri}]{tran2015learning}
Du~Tran, Lubomir Bourdev, Rob Fergus, Lorenzo Torresani, and Manohar Paluri.
  2015.
\newblock Learning spatiotemporal features with 3d convolutional networks.
\newblock In \emph{Proceedings of the IEEE International Conference on Computer
  Vision}, pages 4489--4497.

\bibitem[{Tran et~al.(2018)Tran, Wang, Torresani, Ray, LeCun, and
  Paluri}]{tran2018closer}
Du~Tran, Heng Wang, Lorenzo Torresani, Jamie Ray, Yann LeCun, and Manohar
  Paluri. 2018.
\newblock A closer look at spatiotemporal convolutions for action recognition.
\newblock In \emph{Proceedings of the IEEE Conference on Computer Vision and
  Pattern Recognition (CVPR)}, pages 6450--6459.

\bibitem[{Vaswani et~al.(2017)Vaswani, Shazeer, Parmar, Uszkoreit, Jones,
  Gomez, Kaiser, and Polosukhin}]{vaswani2017attention}
Ashish Vaswani, Noam Shazeer, Niki Parmar, Jakob Uszkoreit, Llion Jones,
  Aidan~N Gomez, {\L}ukasz Kaiser, and Illia Polosukhin. 2017.
\newblock Attention is all you need.
\newblock In \emph{Advances in neural information processing systems}, pages
  5998--6008.

\bibitem[{Yu et~al.(2019)Yu, Yu, Cui, Tao, and Tian}]{yu2019deep}
Zhou Yu, Jun Yu, Yuhao Cui, Dacheng Tao, and Qi~Tian. 2019.
\newblock Deep modular co-attention networks for visual question answering.
\newblock In \emph{Proceedings of the IEEE Conference on Computer Vision and
  Pattern Recognition}, pages 6281--6290.

\bibitem[{Zadeh et~al.(2018{\natexlab{a}})Zadeh, Liang, Mazumder, Poria,
  Cambria, and Morency}]{zadeh2018memory}
Amir Zadeh, Paul~Pu Liang, Navonil Mazumder, Soujanya Poria, Erik Cambria, and
  Louis-Philippe Morency. 2018{\natexlab{a}}.
\newblock Memory fusion network for multi-view sequential learning.
\newblock In \emph{Thirty-Second AAAI Conference on Artificial Intelligence}.

\bibitem[{Zadeh et~al.(2018{\natexlab{b}})Zadeh, Liang, Poria, Cambria, and
  Morency}]{bagher-zadeh-etal-2018-multimodal}
AmirAli Zadeh, Paul~Pu Liang, Soujanya Poria, Erik Cambria, and Louis-Philippe
  Morency. 2018{\natexlab{b}}.
\newblock \href {https://doi.org/10.18653/v1/P18-1208} {Multimodal language
  analysis in the wild: {CMU}-{MOSEI} dataset and interpretable dynamic fusion
  graph}.
\newblock In \emph{Proceedings of the 56th Annual Meeting of the Association
  for Computational Linguistics (Volume 1: Long Papers)}, pages 2236--2246,
  Melbourne, Australia. Association for Computational Linguistics.

\end{thebibliography}
\bibliographystyle{acl_natbib}

\end{document}